\documentclass{article}

\usepackage{arxiv}

\usepackage[utf8]{inputenc} 
\usepackage[T1]{fontenc}    
\usepackage{hyperref}       
\usepackage{url}            
\usepackage{booktabs}       
\usepackage{amsfonts}       
\usepackage{nicefrac}       
\usepackage{microtype}      
\usepackage{lipsum}
\usepackage{graphicx}
\graphicspath{ {./images/} }

\usepackage{amsmath,amssymb,amsfonts}
\usepackage{textcomp}
\usepackage{xcolor}
\usepackage{array}
\usepackage{multirow}
\usepackage{algorithm}
\usepackage{algpseudocode}
\usepackage{cite}
\usepackage{soul}
\usepackage{caption}
\usepackage{booktabs}
\usepackage{array}
\usepackage{comment}
\usepackage{amsmath}
\usepackage{amssymb}
\usepackage{mathtools}
\usepackage{stfloats}
\usepackage{colortbl}
\usepackage{multirow}
\usepackage{siunitx}
\usepackage{csquotes}
\usepackage{subfig}
\usepackage{caption}
\usepackage{subcaption}
\usepackage{pifont}
\usepackage{ulem,xpatch}
\xpatchcmd{\sout}
  {\bgroup}
  {\bgroup}
  {}{}
\usepackage{graphicx}
\usepackage{booktabs}
\usepackage{array}
\usepackage{multirow}
\usepackage{subfig}
\usepackage{caption}
\usepackage{comment}
\usepackage{soul}
\usepackage{multirow}
\usepackage{booktabs} 
\xpatchcmd{\sout} 
{\bgroup}
{\bgroup}
{}{}
\newcolumntype{M}[1]{>{\centering\arraybackslash}m{#1}}

\def\BibTeX{{\rm B\kern-.05em{\sc i\kern-.025em b}\kern-.08em
    T\kern-.1667em\lower.7ex\hbox{E}\kern-.125emX}}

\title{Parameter-Efficient Adaptation of mPLUG-Owl2 via Pixel-Level Visual Prompts for NR-IQA}

\author{
\textbf{Yahya Benmahane} \\
  Computer Science Department\\
  Faculty of Sciences, Rabat \\
  \texttt{c.benmahane@gmail.com} \\
  \And
\textbf{Mohammed El Hassouni} \\
  Computer Science Department\\
  FLSH \\
  \texttt{mohamed.elhassouni@flsh.um5.ac.ma} \\
}

\begin{document}
\maketitle

\begin{abstract}
In this paper, we propose a novel parameter-efficient adaptation method for No-Reference Image Quality Assessment (NR-IQA) using visual prompts optimized in pixel-space. Unlike full fine-tuning of Multimodal Large Language Models (MLLMs), our approach trains only $\sim600$K parameters at most ($<0.01\%$ of the base model), while keeping the underlying model fully frozen. During inference, these visual prompts are combined with images via addition and processed by mPLUG-Owl2 with the textual query "Rate the technical quality of the image." Evaluations across distortion types (synthetic, realistic, AI-generated) on KADID-10k, KonIQ-10k, and AGIQA-3k demonstrate competitive performance against full finetuned methods and specialized NR-IQA models, achieving 0.93 SRCC on KADID-10k. To our knowledge, this is the first work to leverage pixel-space visual prompts for NR-IQA, enabling efficient MLLM adaptation for low-level vision tasks. \textit{The source code is publicly available at \url{https://github.com/yahya-ben/mplug2-vp-for-nriqa}}.
\end{abstract}

\section{Introduction}


The task of No-Reference Image Quality Assessment (NR-IQA) aims to evaluate the quality of an image when no reference image is available. Ideally, a human evaluator is able to accomplish this task. While accurate, this approach is also laborious. Automatic image  quality evaluators are devised to solve this task efficiently \cite{mittalMakingCompletelyBlind2013}\cite{zhangBlindImageQuality2020}\cite{keMUSIQMultiscaleImage2021}.

The development of NR-IQA models has been consistently inspired by trends of machine learning research. Its recent inspiration being Multimodal Large Language Models (MLLMs). These models are pretrained on large amounts of data, offering great generalization capability. Additionally, the introduction of another modality, language, adds a new interpretable output on top of the usual numerical quality scores \cite{wuQInstructImprovingLowlevel2023}\cite{wuQAlignTeachingLMMs2023a}.

Numerous works are exploring this new direction. CLIPIQA achieved competitive Mean Opinion Scores (MOS) correlation without task-specific training, marking the promise of these models as quality evaluators \cite{wangExploringCLIPAssessing2022}. Q-Bench established the first benchmark to assess the low-level vision ability of these multimodal models \cite{wuQBenchBenchmarkGeneralPurpose2024}. They too highlight the promise of these models for NR-IQA and suggest efforts to enhance their quality assessment capabilities. Subsequently, works focused on improving this capability in a training-free fashion by handcrafting textual prompts \cite{wuComprehensiveStudyMultimodal2024a}, or, by fully finetuning these models on task-specific datasets \cite{zhangBlindImageQuality2023}\cite{wuQInstructImprovingLowlevel2023}\cite{wuQAlignTeachingLMMs2023a}. However, training-free methods still fall short in terms of performance, and fully finetuning an MLLM guarantees improved performance at a memory cost, and potential model forgetting.

In this paper, we propose the exploration of visual prompting as a parameter-efficient method for adapting mPLUG-Owl2 \cite{yeMPLUGOwl2RevolutionizingMultimodal2023} for the NR-IQA task \cite{bahngExploringVisualPrompts2022}. By learning a set of pixels to combine with the input image while keeping all of mPLUG-Owl2's \cite{yeMPLUGOwl2RevolutionizingMultimodal2023} parameters frozen, we are able to accomplish an efficient and non-invasive solution to promote MLLMs as reliable quality evaluators.

Through our experiments, we show that learning such a visual prompt offers competitive performance on established IQA datasets \cite{hosuKonIQ10kEcologicallyValid2020}\cite{linKADID10kLargescaleArtificially2019}\cite{liAGIQA3KOpenDatabase2023}.

\begin{figure*}[t]
    \centering
    \includegraphics[width=\textwidth]{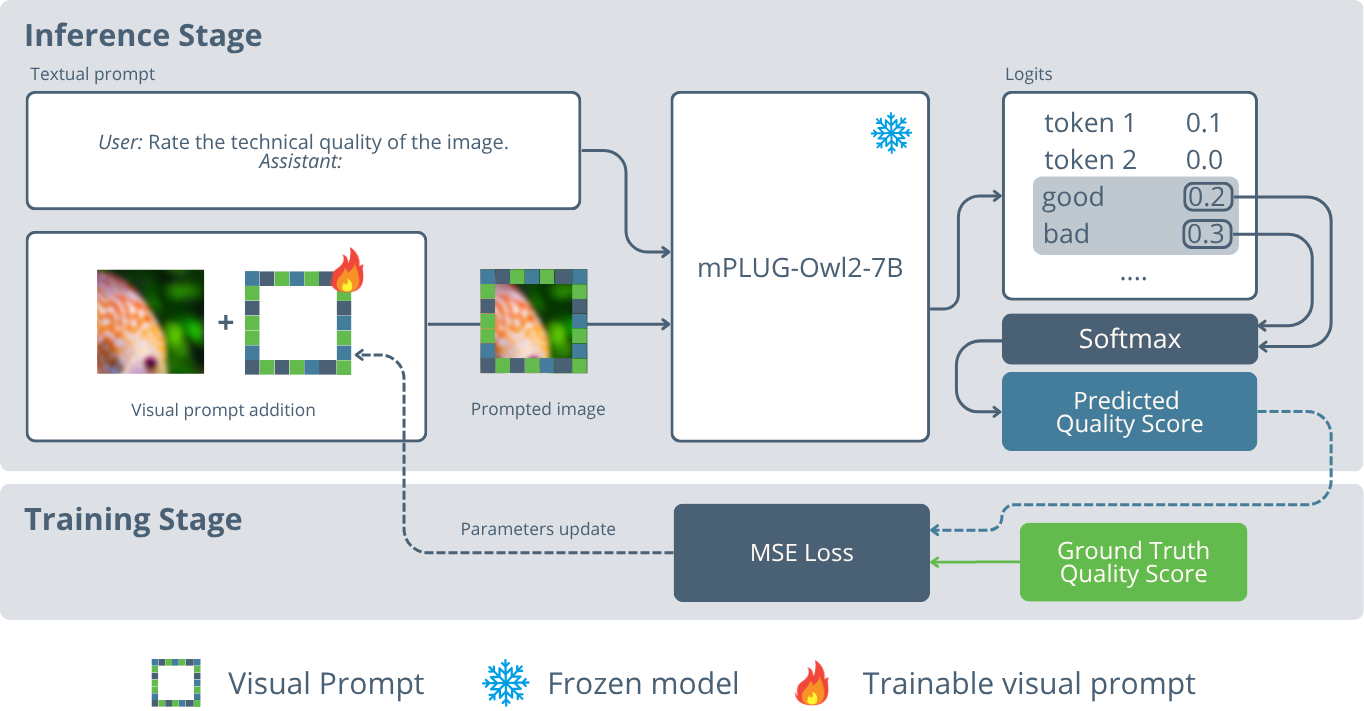}
    \caption{Our proposed method for efficiently adapting mPLUG-Owl2 \cite{yeMPLUGOwl2RevolutionizingMultimodal2023} for NR-IQA using visual prompts.}
    
    \label{fig:wide_image}
\end{figure*}

To the best of our knowledge, we are the first to explore visual prompting for the NR-IQA, a task that is not explored in works focusing on fundamental visual prompting research  \cite{bahngExploringVisualPrompts2022}\cite{wuUnleashingPowerVisual2023}\cite{tsaoAutoVPAutomatedVisual2024}\cite{chenUnderstandingImprovingVisual2023}.

\section{Related Work}

\textbf{No-Reference Image Quality Assessment.} Similar to most computer vision tasks, NR-IQA largely followed a similar trajectory of development. Earlier methods relied on handcrafting quality aware features to learn a quality evaluator using training images with human subjective scores  \cite{mittalMakingCompletelyBlind2013}. The advent of deep learning methods, coupled with the creation of IQA specific datasets covering multiple image distortion types \cite{hosuKonIQ10kEcologicallyValid2020}\cite{linKADID10kLargescaleArtificially2019}, enabled the end-to-end training on sophisticated network architectures, which improved baselines across different distortion types \cite{maEndtoEndBlindImage2018}\cite{zhangBlindImageQuality2020}\cite{zhangUncertaintyAwareBlindImage2021}\cite{keMUSIQMultiscaleImage2021}\cite{yangMANIQAMultidimensionAttention2022}\cite{madhusudanaImageQualityAssessment2021}.

\textbf{MLLMs for No-Reference Image Quality Assessment.} The introduction of MLLMs attracted the attention of the NR-IQA community to harness their capabilities. CLIPIQA first explored the use of CLIP for assessing image quality. Directly prompting CLIP for a quality score showed little correlation with human subjective scores \cite{wangExploringCLIPAssessing2022}. A contrastive prompt pairing strategy of negative and positive prompts achieved better correlation, reaffirming prompt engineering as a key challenge facing the adaptation of MLLMs for new tasks \cite{radfordLearningTransferableVisual2021}. While CLIPIQA demonstrated the promising zero-shot NR-IQA ability of MLLMs, improved performance resulted by using CoOp \cite{zhouLearningPromptVisionLanguage2022a} which introduces trainable components to the textual prompt while keeping the model weights frozen \cite{wangExploringCLIPAssessing2022}, demonstrating the utility of prompt optimizations which we extend to the pixel-space in this work. In a similar vein, LIQE adopted a multitask learning strategy to fully finetune CLIP, achieving improved performance over the pretrained CLIP \cite{zhangBlindImageQuality2023}.

Beyond using CLIP as a quality evaluator, Q-Bench \cite{wuQBenchBenchmarkGeneralPurpose2024} evaluated the low-level vision abilities of multiple MLLMs. Their findings validated the potential of these models for NR-IQA, while calling for further improvements. To evaluate the NR-IQA ability of MLLMs, Q-Bench resorted to a softmax-based strategy to mitigate experimentally observed biases of directly adopting MLLM generated quality scores \cite{wuQBenchBenchmarkGeneralPurpose2024}. To enhance MLLMs performance for NR-IQA and further their overall low-level vision abilities, Q-Instruct \cite{wuQInstructImprovingLowlevel2023} pioneered finetuning open-source multimodal LLMs on a custom dataset. The instruction tuning resulted in state-of-the-art performance surpassing non MLLM-based NR-IQA models. Concurrently, Q-Align finetuned an open-source multimodal to output qualitative quality levels instead of numerical quality scores, mimicking the natural behavior of human evaluators, further improving the performance of MLLM based NR-IQA models \cite{wuQAlignTeachingLMMs2023a}. 

However, fully finetuning a large multimodal model is inefficient, and introduces the risk of model forgetting. Conversely, learning a visual prompt while freezing the multimodal model parameters remains comparatively efficient \cite{bahngExploringVisualPrompts2022}.

\textbf{Visual prompting.} Multimodal LLMs models proved highly sensitive to minor prompt changes, requiring meticulous handcrafted textual prompts \cite{radfordLearningTransferableVisual2021}\cite{wuComprehensiveStudyMultimodal2024a}. Efforts are being made for the automatic optimization of prompts to adapt models to new tasks and boost their performance \cite{liSurveyAutomaticPrompt2025}. The general framework of prompt optimization is the adaptation of the input space while keeping the model's parameters frozen. In the case of a vision-language model, both the textual prompt and visual prompt can be optimized. In this work, we aim to optimize the latter.

Prior work investigated visual prompting for CLIP and for vision models \cite{bahngExploringVisualPrompts2022}. Competitive performance resulted on several image classification datasets in comparison to the full finetuning or linear probing of these same models. Follow-up works focused on further improving visual prompting \cite{wuUnleashingPowerVisual2023}\cite{tsaoAutoVPAutomatedVisual2024}, learning the visual prompt with no access to the model's parameters \cite{ohRobustAdaptationFoundation2024} and exploring important aspects of visual prompting like label mapping \cite{chenUnderstandingImprovingVisual2023}. We are inspired by these works, but acknowledge the missing extension of visual prompting to the NR-IQA task, a gap we aim to fill in this work. 

It is worth mentioning that a flavor of visual prompting has previously been explored for NR-IQA \cite{luMultiLayerCrossModalPrompt2025}\cite{panMultiModalPromptLearning2024}\cite{luQAdaptAdaptingLMM2025}. However, our work is different because no introduction of learnable components into the multimodal model's architecture takes place, our goal is the pixel-space adaptation of mPLUG-Owl2 \cite{yeMPLUGOwl2RevolutionizingMultimodal2023} for NR-IQA.

\begin{figure*}[t]
    \centering
    \includegraphics[width=1.0\linewidth]{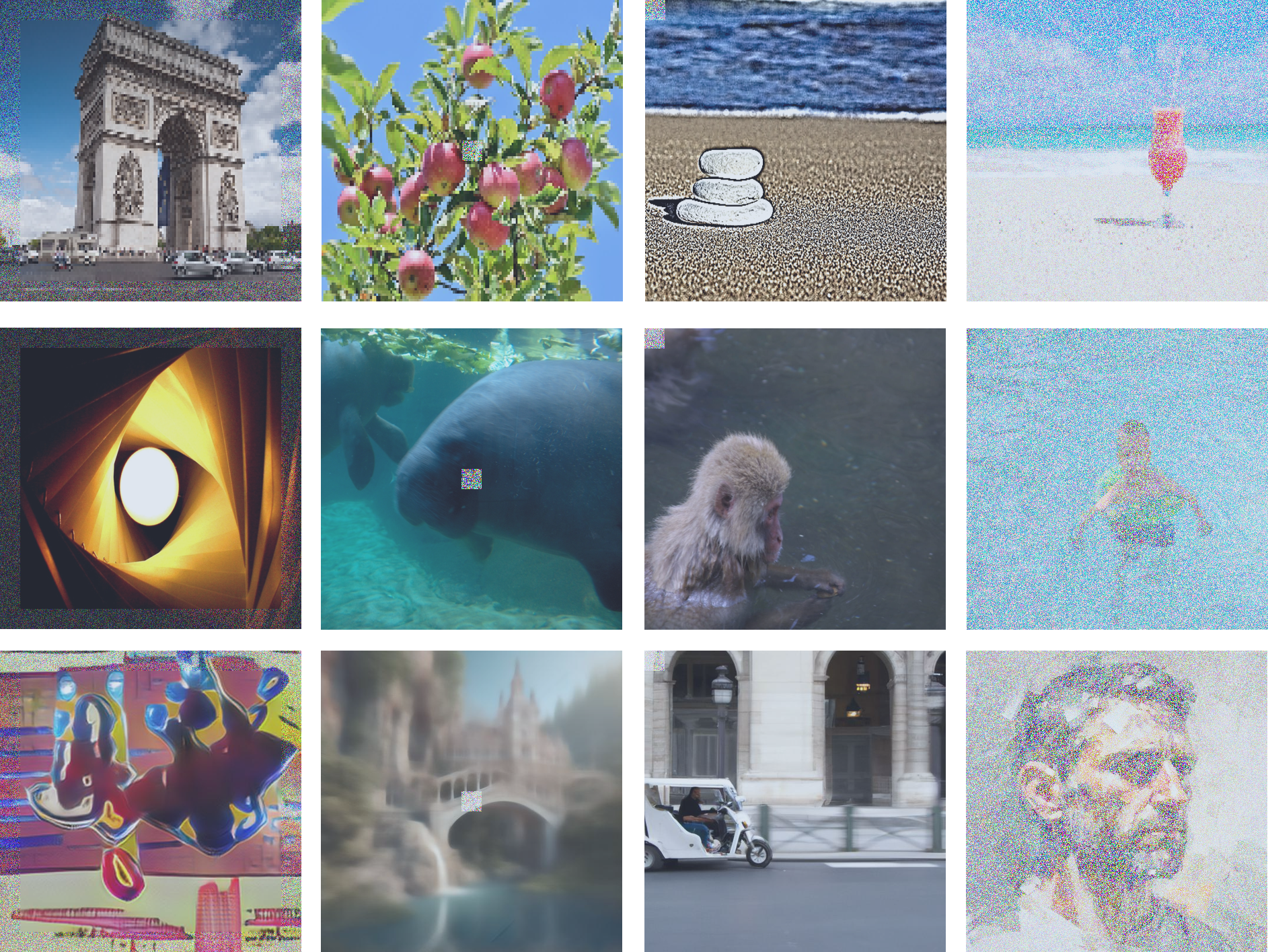}
    \caption{The following figure illustrates four types of visual prompts applied to sample images from three datasets (top to bottom: KADID-10k \cite{linKADID10kLargescaleArtificially2019}, KonIQ-10k \cite{hosuKonIQ10kEcologicallyValid2020}, AGIQA-3k \cite{liAGIQA3KOpenDatabase2023}). The pixelated regions represent the learned visual prompts: (a) The first column shows a 30px padding surrounding the image on all sides. (b) The second column displays a 30px square patch centered in the image. (c) The third column presents a 30px square patch located at the top-left corner of the image. (d) The fourth column shows a visual prompt applied as a full overlay across the entire image.}
    \label{fig:visual_prompts}
\end{figure*}

\section{Method}

To adapt mPLUG-Owl2 \cite{yeMPLUGOwl2RevolutionizingMultimodal2023} for the NR-IQA task in a parameter-efficient way, we aim to learn a visual prompt to apply to the input image. During inference, we feed a textual prompt along with a visually prompted image to the frozen MLLM. The prompted image is a combination of a trainable visual prompt and an input image. The output logits for quality-related tokens are passed to a softmax function to compute a quality score. During training, only the visual prompt is updated. 

In this section, we first present the overall pipeline of our method (see in Figure \ref{fig:wide_image}). Then we describe the core components of the proposed method: \textbf{(a)} Choice of the MLLM, \textbf{(b)} Visual prompts and \textbf{(c)} textual prompts configurations.

\subsection{Overall Pipeline}

Consider a training set \(
  \mathcal{D} = \bigl\{(x_i,\,y_i)\bigr\}_{i=1}^N,
\) where each \(x_i\) is an input image and \(y_i \in [0,1]\) is its ground‐truth quality score. \(N\) denotes the number of training samples. Feeding the prompted image \(x_i'\) together with a fixed textual prompt \(t\) into a frozen MLLM (with parameters \(\theta\)) yields a sequence of token‐logits at positions \(1 \leq t \leq T\):
\[
  \bigl[\ell_{i,1},\,\ell_{i,2},\,\dots,\,\ell_{i,T}\bigr]
  \;=\;
  \mathrm{MLLM}_{\theta}\bigl(x_i',\,t;\,\theta\bigr),
  \quad
  \ell_{i,t}\in\mathbb{R}^V,
\]
where \(V\) denotes the vocabulary size and \(T\) is the sequence length.  We retain only the final logit vector which corresponds to the model's response
\[
  \ell_{i,T} \in \mathbb{R}^V.
\]

To extract a single "quality" score from \(\ell_{i,T}\), we introduce two disjoint sets of token‐IDs:
\[
  P = \{\text{positive token IDs}\}, 
  \qquad
  N = \{\text{negative token IDs}\}.
\]
For instance, if "good" has ID \(100\in P\) and "bad" has ID \(200\in N\), then \(\ell_{i,T}^{\,100}\) and \(\ell_{i,T}^{\,200}\) are the corresponding logits.  In general, the logits for all positives are \(\{\ell_{i,T}^j : j\in P\}\), and for negatives \(\{\ell_{i,T}^k : k\in N\}\). Both the positive and negative token-IDs sets are populated based on mPLUG-Owl2's \cite{yeMPLUGOwl2RevolutionizingMultimodal2023} specific vocabulary.

A scalar quality score \(s_i \in (0,1)\) is defined by a softmax-like calculation:
\[
  s_i 
  = 
  \frac{\displaystyle \sum_{j \in P} \exp\bigl(\ell_{i,T}^j\bigr)}
       {\displaystyle \sum_{j \in P} \exp\bigl(\ell_{i,T}^j\bigr)
       + \sum_{k \in N} \exp\bigl(\ell_{i,T}^k\bigr)}.
\]

We favor this approach instead of either prompting the model for a quality related class and then mapping back to a numerical quality score \cite{wuQAlignTeachingLMMs2023a}, or, asking the model to directly output a quality score \cite{wuComprehensiveStudyMultimodal2024a}. We believe in the same empirical results demonstrated in \cite{wuQBenchBenchmarkGeneralPurpose2024}.

We formulate the NR-IQA task as a regression problem and optimize a visual prompt via backpropagation by minimizing the Mean Squared Error (MSE). Comparing \(s_i\) to the ground truth \(y_i\) via mean‐squared error gives
\[
  \mathcal{L}_i = \bigl(s_i - y_i\bigr)^2,
\]
and over the entire training set,
\[
  \mathcal{L}(p)
  = \frac{1}{N}\sum_{i=1}^N \bigl(s_i - y_i\bigr)^2.
\]

Thus, the optimization problem becomes finding the visual prompt \(p^*\) that minimizes \(\mathcal{L}(p)\):

\[
  p^* \;=\; \arg\min_{p} \frac{1}{N}\sum_{i=1}^N \bigl(s_i(p) - y_i\bigr)^2
\]

\subsection{Multimodal LLM selection}

We are interested in this work to use multimodal LLMs to train a visual prompt. Essentially, there is a wide choice of MLLMs to experiment with \cite{yinSurveyMultimodalLarge2024}. However, we found that few MLLMs provide a compatible interface to train a visual prompt. To avoid any alteration of the MLLM internals that might negatively impact the expected format of its vision encoder, we select mPLUG-Owl2-7B \cite{yeMPLUGOwl2RevolutionizingMultimodal2023} for their compatible out-of-the-box API with our current training setup. Additionally, we adopt this model for its strong overall performance \cite{yeMPLUGOwl2RevolutionizingMultimodal2023}. We emphasize that we are keeping the model's parameters frozen and only learning the visual prompt.

\subsection{Visual prompt}

We learn a visual prompt \(p\) to apply to an image \(x\). A visual prompt can take various shapes and sizes, and can be located anywhere on the image. Table~\ref{tab:prompt_summary} summarizes the proposed visual prompts. Inspired by previous work on visual prompting \cite{bahngExploringVisualPrompts2022}, we consider padding and fixed patches. Additionally, we introduce the full overlay as an extension of the fixed patch that covers the entire image. 
\begin{table}[h]
\centering
\caption{Summary of visual prompt configurations.}
\resizebox{\columnwidth}{!}{
\begin{tabular}{llll}
\toprule
\textbf{Visual prompt type} & \textbf{Size(s)} & \textbf{Location} & \textbf{\# Trainable parameters} \\
\midrule
Padding         & $S=10$, $30$     & Around all image borders & $3 \times 2S(W + H - S)$ \\
Fixed Patch (Center)    & $S=10$, $30$     & Center of the image       & $3 \times S^2$ \\
Fixed Patch (Top-Left)  & $S=10$, $30$     & Top-left corner           & $3 \times S^2$ \\
Full Overlay    & Full image       & Entire image surface      & $3 \times H \times W$ \\
\bottomrule
\end{tabular}
}
\label{tab:prompt_summary}
\end{table}
Regarding size, we experiment with 30px and 10px. For location, the fixed patches are tested at the center of the image and at the top-left corner (Figure~\ref{fig:visual_prompts}). Consistent with prior work \cite{bahngExploringVisualPrompts2022}, we add the visual prompt directly to the input image. Applying the visual prompt \(p\) to an image \(x\) results in a prompted image \(x'\): 
\[
  x' = x + p\\
\]
It is worth mentioning that we also introduce restricting the pixel values of the visual prompts with a $\tanh$ function to a range of $[-1, 1]$ and then clamp the resulting pixel values to a range of $[0, 1]$.

As shown in Table~\ref{tab:prompt_summary}, the number of trainable parameters varies across prompt types. For an image of size \(C \times H \times W\) (e.g., \(3 \times 448 \times 448\)), a full overlay contains \(3 \times 448 \times 448 = 602{,}112\) parameters, while a padding of size 30px includes \(3 \times 2 \times 30 \times (448 + 448 - 30) = 155{,}880\) parameters. A fixed patch of size 30px contains \(3 \times 30^2 = 2{,}700\) parameters, and a 10px patch contains \(3 \times 10^2 = 300\) parameters. Despite the number of parameters, this remains comparatively modest relative to full model fine-tuning.

\subsection{Textual prompt}

Recent work investigated label mapping as a key factor in visual prompting \cite{chenUnderstandingImprovingVisual2023}. In the case of vision-language models, the textual prompt plays the same critical role. We select the same textual prompt used in Q-Bench for quality assessment \cite{wuQBenchBenchmarkGeneralPurpose2024}, with the slight but important addition of the word ``technical". We believe that using the prompt \textit{"Rate the quality of the image."} alone with no addition of the word "technical" is ambiguous. Other variations of textual prompts are possible. When training our visual prompts, we stick to the textual prompt outlined just before. 

The positive tokens extracted from the final logits vector are ``good'' and ``fine'', whereas the negative tokens are ``poor'' and ``bad''.

\subsection{Results}
\begin{table*}[t]
\begin{minipage}{\textwidth}
\caption{SRCC and PLCC performance of different visual prompt types across three IQA datasets for mPLUG-Owl2 \cite{yeMPLUGOwl2RevolutionizingMultimodal2023}.}
\label{tab:tab_2}
\small
\setlength{\tabcolsep}{4pt}
\renewcommand{\arraystretch}{0.8}
\resizebox{\textwidth}{!}{
\begin{tabular}{ll|cc|cc|cc}
\toprule
\multicolumn{2}{c|}{\textbf{Prompt Type}} & 
\multicolumn{2}{c|}{\textbf{KADID-10k \cite{linKADID10kLargescaleArtificially2019}}} & 
\multicolumn{2}{c|}{\textbf{KonIQ-10k \cite{hosuKonIQ10kEcologicallyValid2020}}} & 
\multicolumn{2}{c}{\textbf{AGIQA-3k \cite{liAGIQA3KOpenDatabase2023}}} \\
\cmidrule{3-8}
\textbf{Type} & \textbf{Size} & SRCC & PLCC & SRCC & PLCC & SRCC & PLCC \\
\midrule
\multirow{2}{*}{Padding} 
  & 10px & 0.880 & 0.873 & 0.805 & 0.840 & 0.802 & 0.851 \\
  & 30px & \textbf{0.932} & \textbf{0.929} & \textbf{0.852} & \textbf{0.874} & \textbf{0.810} & \textbf{0.860} \\
\midrule
\multirow{2}{*}{Fixed Patch (Center)} 
  & 10px & 0.390 & 0.344 & 0.487 & 0.415 & 0.435 & 0.397 \\
  & 30px & 0.806 & 0.780 & 0.647 & 0.687 & 0.725 & 0.781 \\
\midrule
\multirow{2}{*}{Fixed Patch (Top-Left)} 
  & 10px & 0.465 & 0.438 & 0.551 & 0.518 & 0.564 & 0.530 \\
  & 30px & 0.520 & 0.528 & 0.635 & 0.691 & 0.755 & 0.794 \\
\midrule
Full Overlay & — & 0.887 & 0.893 & 0.693 & 0.726 & 0.624 & 0.694 \\
\bottomrule
\end{tabular}
}
\end{minipage}
\end{table*}

\section{Experiments}

\subsection{Experimental setup}

To evaluate the effectiveness of visual prompting for adapting mPLUG-Owl2 \cite{yeMPLUGOwl2RevolutionizingMultimodal2023} for the NR-IQA task, we conduct experiments on three IQA datasets covering different image distortion types. KADID-10k \cite{linKADID10kLargescaleArtificially2019} for synthetic distortions: contains a total of 81 high-quality reference images subject to 25 artificial distortions along 5 levels of degradation, resulting in 10{,}125 distorted images. KonIQ-10k \cite{hosuKonIQ10kEcologicallyValid2020} for realistic distortions: this dataset contains 10k images collected from the internet. The dataset is diverse in terms of image quality and content. AGIQA-3k \cite{liAGIQA3KOpenDatabase2023} for AI-generated images: with the advent of image generators, the performance of these models depends on the quality of the generated images. Developing image quality metrics to assess the performance of these models along this axis is needed. AGIQA-3k \cite{liAGIQA3KOpenDatabase2023} contains 3k diverse AI-generated images using different image generation models.

We use the official train/test splits when available \cite{hosuKonIQ10kEcologicallyValid2020}. Otherwise, we select 80\% of the dataset for training, 10\% for validation, and the remaining 10\% for testing. For AGIQA-3k \cite{liAGIQA3KOpenDatabase2023}, due to its limited size, 60\% of the dataset is used for training and 20\% is respectively used for validation and testing.

\subsection{Implementation details}
We implement our training and inference loops in PyTorch. We load the multimodal LLM with no quantization. We choose the 7B variant of mPLUG-Owl2 \cite{yeMPLUGOwl2RevolutionizingMultimodal2023}. Due to the MLLM's default image size, we resize and center crop the input image to fit the expected format of the MLLM. We use Stochastic Gradient Descent (SGD) as optimizer. We apply random horizontal flipping for data augmentation. We also apply the expected normalization \textbf{after} applying the visual prompt to the image and before feeding the image to the MLLM, which we experimentally verified to yield superior performance compared to omitting normalization or applying it prior to the visual prompt. We train the visual prompts on KADID-10k \cite{linKADID10kLargescaleArtificially2019} with a batch size of 32, a learning rate of 60, and for 25 epochs. For the visual prompt padding (30px), we extend training by an additional 25 epochs with a reduced learning rate of 20. For KonIQ-10k \cite{hosuKonIQ10kEcologicallyValid2020} and AGIQA-3k \cite{liAGIQA3KOpenDatabase2023}, the batch size is set to 4, with AGIQA-3k trained for an additional 10 epochs. All experiments are conducted on a single NVIDIA RTX A6000 GPU.

\begin{table*}[t]
\begin{minipage}{\textwidth}
\caption{Performance comparison of our method on KADID-10k 
\cite{linKADID10kLargescaleArtificially2019}, KonIQ-10k \cite{hosuKonIQ10kEcologicallyValid2020} and AGIQA-3k \cite{liAGIQA3KOpenDatabase2023} datasets.}
\label{tab:tab_3}
\small
\setlength{\tabcolsep}{4pt}
\renewcommand{\arraystretch}{0.8}
\resizebox{\textwidth}{!}{
\begin{tabular}{l|cc|cc|cc}
\toprule
\textbf{Methods} & 
\multicolumn{2}{c|}{\textbf{KADID-10k \cite{linKADID10kLargescaleArtificially2019}}} & 
\multicolumn{2}{c|}{\textbf{KonIQ-10k \cite{hosuKonIQ10kEcologicallyValid2020}}} & 
\multicolumn{2}{c}{\textbf{AGIQA-3k \cite{liAGIQA3KOpenDatabase2023}}} \\
  & SRCC$\uparrow$ & PLCC$\uparrow$ & SRCC$\uparrow$ & PLCC$\uparrow$ & SRCC$\uparrow$ & PLCC$\uparrow$ \\
\midrule
\multicolumn{7}{l}{\textbf{Specialized Models}} \\
\midrule
DBCNN \cite{zhangBlindImageQuality2020}        & 0.878 & 0.878 & 0.864 & 0.868 & -     & -     \\
HyperIQA \cite{suBlindlyAssessImage2020}        & 0.872 & 0.869 & 0.906 & 0.917 & -     & -     \\
TreS \cite{golestanehNoReferenceImageQuality2022}         & 0.858 & 0.859 & 0.928 & 0.915 & -     & -     \\
UNIQUE \cite{zhangUncertaintyAwareBlindImage2021}         & 0.878 & 0.876 & 0.896 & 0.901 & -     & -     \\
MUSIQ \cite{keMUSIQMultiscaleImage2021}         & -     & -     & 0.916 & 0.928 & -     & -     \\
\midrule
\multicolumn{7}{l}{\textbf{Training-free}} \\
\midrule
mPLUG-Owl2-7b\cite{yeMPLUGOwl2RevolutionizingMultimodal2023} & 0.550 & 0.571 & 0.602 & 0.677 & 0.436 & 0.422 \\
\midrule
\multicolumn{7}{l}{\textbf{CoOp \cite{zhouLearningPromptVisionLanguage2022a}}} \\
\midrule
CLIPIQA+ \cite{wangExploringCLIPAssessing2022}  & -     & -     & 0.895 & 0.909 & -     & -     \\
\midrule
\multicolumn{7}{l}{\textbf{Integrated Visual Prompting}} \\
\midrule
MP-IQE \cite{panMultiModalPromptLearning2024}  & 0.941     & 0.944     & 0.898 & 0.904 & -     & -     \\
MCPF-IQA \cite{luMultiLayerCrossModalPrompt2025}  & -     & -     & 0.918 & 0.919 &  0.872     & 0.919    \\
Q-Adapt \cite{luQAdaptAdaptingLMM2025}  & 0.769 & 0.754     & 0.878 & 0.907 & 0.757 & 0.789      \\
\midrule
\multicolumn{7}{l}{\textbf{Fully Finetuned}} \\
\midrule
LIQE \cite{zhangBlindImageQuality2023}          & 0.930 & 0.931 & 0.919 & 0.908 & -     & -     \\
Q-Align \cite{wuQAlignTeachingLMMs2023a}        & 0.919 & 0.918 & 0.940 & 0.941 & 0.727 & 0.795 \\
Q-Instruct \cite{wuQInstructImprovingLowlevel2023} & 0.706 & 0.710 & 0.911 & 0.921 & 0.772 & 0.847 \\
\midrule
\multicolumn{7}{l}{\textbf{Ours}} \\
\midrule
\textbf{Our Proposed Method} & \textbf{0.932} & \textbf{0.929} & \textbf{0.852} & \textbf{0.865} & \textbf{0.810} & \textbf{0.860} \\
\bottomrule
\end{tabular}
}
\end{minipage}
\end{table*}

Table \ref{tab:tab_2} shows the performance of four learned visual prompts (Padding, Fixed Patch (Center), Fixed Patch (Top-Left) and Full Overlay) across three IQA datasets (KADID-10k \cite{linKADID10kLargescaleArtificially2019}, KonIQ-10k \cite{hosuKonIQ10kEcologicallyValid2020}, AGIQA-3k \cite{liAGIQA3KOpenDatabase2023}) on mPLUG-Owl2-7B \cite{yeMPLUGOwl2RevolutionizingMultimodal2023}. We downsize the visual prompts to 10px (except the full overlay) and train an additional variant of each visual prompt to test the effect of prompt size. Center and Top-Left denote the position of the patch in the image. Performance is evaluated with SRCC and PLCC.

Based on these results, we notice that the 30px padding is the most correlated type of visual prompt with the MOS across all datasets. The newly introduced full overlay shows good correlation on the KADID-10k dataset \cite{linKADID10kLargescaleArtificially2019}. Other variants of visual prompts show little correlation with subjective scores. We also notice a drop in performance when using small size visual prompts. For example, downsizing from a 30px padding to a 10px degrades performance. The use of 10px fixed patches also degrades performance regardless of its position.

The effectiveness of visual prompting in adapting mPLUG-Owl2-7B \cite{yeMPLUGOwl2RevolutionizingMultimodal2023} is highlighted in Table \ref{tab:tab_3}. We evaluate on three datasets using our inference pipeline without visual prompts and compare it against our method, which incorporates lightweight pixel-level visual prompts. Our proposed method outperforms the pretrained-only mPLUG-Owl2-7B baseline.

We compare our method against fully finetuned MLLMs on the IQA task, including Q-Instruct \cite{wuQInstructImprovingLowlevel2023}, Q-Align \cite{wuQAlignTeachingLMMs2023a}, and LIQE \cite{zhangBlindImageQuality2023}. Q-Instruct was trained on a mix of datasets 95.26\% SPAQ \cite{fangPerceptualQualityAssessment2020}, 48.92\% KonIQ-10k \cite{hosuKonIQ10kEcologicallyValid2020}, 2\% LIVE-FB \cite{yingPatchesPicturesPaQ2PiQ2019a}, 17.11\% LIVE-itw \cite{MassiveOnlineCrowdsourced}, and 13.41\% AGIQA-3k \cite{liAGIQA3KOpenDatabase2023} but not on KADID-10k \cite{linKADID10kLargescaleArtificially2019}. We report its best-performing results. Q-Align \cite{wuQAlignTeachingLMMs2023a} was trained and tested on KonIQ-10k \cite{hosuKonIQ10kEcologicallyValid2020} and KADID-10k \cite{linKADID10kLargescaleArtificially2019}, but not on AGIQA-3k \cite{liAGIQA3KOpenDatabase2023}. For AGIQA-3k \cite{liAGIQA3KOpenDatabase2023}, we report results from its training on KonIQ-10k \cite{hosuKonIQ10kEcologicallyValid2020} combined with SPAQ \cite{fangPerceptualQualityAssessment2020}. LIQE \cite{zhangBlindImageQuality2023} finetunes two CLIP-based variants using KADID-10k and KonIQ-10k with a ViT-B/32 image encoder and a 63M parameter GPT-2 text encoder.

Our proposed method shows competitive performance compared to these finetuned approaches at a fraction of their memory and storage costs. Both Q-Instruct \cite{wuQInstructImprovingLowlevel2023} and Q-Align \cite{wuQAlignTeachingLMMs2023a} full finetune a 7B multimodal LLMs. Additionally, instead of storing separate models trained on each dataset, our method leverages a single base model adapted through lightweight pixel-level visual prompts. The observed performance gap with some fully finetuned methods may be mitigated by using aggregated training datasets, as done in \cite{wuQInstructImprovingLowlevel2023}, or by extending training into multitask prompt tuning setups \cite{zhangBlindImageQuality2023}.

We observe a notable performance gap with CLIPIQA+, a CoOp-based variant of CLIP finetuned for IQA \cite{wangExploringCLIPAssessing2022}, especially on KonIQ-10k. This highlights the benefits of tuning the textual prompt, in contrast to our method which leaves it untouched.

We also compare our method against models that integrate visual prompts throughout the model architecture. In contrast to our approach, which applies prompting solely at the pixel level, these models optimize visual prompts across multiple layers of the model. Despite this, our pixel-level prompting still shows competitive results. On the KADID-10k dataset, our method surpasses Q-Adapt \cite{luQAdaptAdaptingLMM2025}, a further finetuned version of Q-Instruct, demonstrating the efficiency of pixel-level prompting. However, our method exhibits a performance gap relative to MP-IQE \cite{panMultiModalPromptLearning2024} and MCPF-IQA \cite{luMultiLayerCrossModalPrompt2025}, underlining that deeper visual prompt integration and co-tuning with textual components may provide additional gains.

To assess the adaptability of the multimodal LLM to new tasks without task-specific model design, we also compare against specialized NR-IQA models: MUSIQ \cite{keMUSIQMultiscaleImage2021}, UNIQUE \cite{zhangUncertaintyAwareBlindImage2021}, TreS \cite{golestanehNoReferenceImageQuality2022}, HyperIQA \cite{suBlindlyAssessImage2020}, and DBCNN \cite{zhangBlindImageQuality2020}. For UNIQUE, TreS, HyperIQA, and DBCNN, we report scores from their original papers when available, and otherwise from the LIQE paper \cite{zhangBlindImageQuality2023}. Our method exhibits superior performance on KADID-10k, a historically challenging benchmark, where it consistently achieves the best results. On KonIQ-10k, our method remains competitive but shows a moderate performance gap.

\section{Conclusion}

In this paper, we explore visual prompting for adapting mPLUG-Owl2-7B \cite{yeMPLUGOwl2RevolutionizingMultimodal2023} for the NR-IQA task. By adding a set of learned pixels to the input image, we are able to steer an MLLM to evaluate image quality. Through our experiments, we validated this approach against full finetuned models and specialized IQA models, and demonstrated its competitive performance at a fractional parameter count. This work also extends visual prompting to MLLMs. Future work will focus on better understanding and adapting visual prompting for NR-IQA, extensive hyperparameter tuning, and extending visual prompting to improve other aspects of low-level vision ability of MLLMs.

\bibliographystyle{unsrturl}
\bibliography{main}

\end{document}